\documentclass[letterpaper, 10 pt, conference]{ieeeconf}  % Comment this line out
\IEEEoverridecommandlockouts                              % This command is only
\usepackage[utf8]{inputenc}
\usepackage[T1]{fontenc}
\usepackage{tikz}
\usepackage{stfloats}
\usepackage{xcolor}
\usepackage{booktabs}
\usepackage{amsfonts}
\usepackage{tabularx}
\usepackage{array}
\let\labelindent\relax % ieeeconf already defines it; let enumitem own it
\usepackage{enumitem}
\usepackage{amsmath}
\usepackage{ragged2e}
\usepackage{subcaption}
\usepackage{url}

\usepackage{verbatim}

\newcolumntype{Y}{>{\raggedright\arraybackslash}X}
\newcolumntype{M}[1]{>{\centering\arraybackslash}m{#1}}
\newcolumntype{C}[1]{>{\Centering\arraybackslash}m{#1}}
\newcolumntype{L}[1]{>{\RaggedRight\arraybackslash}m{#1}}

\title{\LARGE \bf
A Minimal Optical-Flow Representation for Vision-Based Tactile Rotation Classification in Robotic Manipulation \\ Across Gravity Domains
}

\author{Oscar Martinez-Bernal$^{1}$, Mario Cavero-Vidal$^{1,2}$, Francesco Grella$^{1}$ and Carol Martinez$^{1}$% <-this % stops a space
\thanks{$^{1}$Space Robotics Research Group (SpaceR), Interdisciplinary Centre for Security, Reliability and Trust (SnT), University of Luxembourg
        {\tt\small  oscar.martinezbernal@uni.lu}}
\thanks{$^{2}$Eurecat, Centre Tecnològic de Catalunya, Robotics and Automation Unit, Barcelona, Spain}
}

\begin{document}

\maketitle
\thispagestyle{empty}
\pagestyle{empty}

%%%%%%%%%%%%%%%%%%%%%%%%%%%%%%%%%%%%%%%%%%%%%%%%%%%%%%%%%%%%%%%%%%%%%%%%%%%%%%%%
\begin{abstract}
Vision-based tactile sensors provide rich contact information, but processing high-resolution images can be costly for resource-constrained platforms such as space robots. 
This work investigates whether a compact representation of tactile motion can classify object rotation across different gravity conditions. Dense optical flow from a simulated GelSight Mini is aggregated over a $7\times9$ grid into 126 features and used to classify the direction of load-induced rotation under Earth, Mars, Moon, and orbital gravity. Gravity causes a small but significant shift in these features, accounting for 1.6\% of their variance ($R^2=0.016$). Despite its small magnitude, this shift affects models trained only on Earth data: XGBoost accuracy decreases from 94.4\% on Earth to 75.9\% in orbit. In contrast, a single model trained across all four gravity domains achieves 96.3\% overall accuracy and 95.1--97.0\% across individual domains, without using gravity as an input. The representation can also be reduced to 40 features while retaining 95.7\% accuracy, with XGBoost requiring only 0.14\,ms per inference. These findings show that Earth-gravity performance alone is insufficient to establish the transferability of tactile perception for space robotic manipulation, highlighting the need to account for gravity-induced domain shifts during training and validation.
\end{abstract}


%%%%%%%%%%%%%%%%%%%%%%%%%%%%%%%%%%%%%%%%%%%%%%%%%%%%%%%%%%%%%%%%%%%%%%%%%%%%%%%%
\section{Introduction}

Contact-rich manipulation is essential for space robotic operations such as on-orbit servicing \cite{Alizadeh2024OOSSurvey}, in-space assembly, maintenance, inspection, and planetary sample collection \cite{Papadopoulos2021SpaceSurvey}. These operations require robots to interpret local contact under limited sensing, uncertain object properties, and gravity conditions that differ from those on Earth. Reliable contact perception is therefore important for safe robotic manipulation \cite{Jahanshahi2025SpaceTactile}. Vision-based tactile sensors offer a promising solution by converting local contact interactions into high-resolution images \cite{Yuan2017GelSight}.

While these images provide rich contact information, processing them directly can be computationally expensive for resource-constrained platforms such as space robots. Compact representations can reduce this computational burden. In particular, optical flow captures tactile motion using a reduced set of features and has been used to estimate displacement, deformation, geometry, and force, as well as for lightweight classification tasks \cite{Zhang2023GelFlow}.

However, tactile perception methods are typically developed and validated under Earth gravity. Changes in gravity affect object dynamics and may therefore alter the tactile signals observed during manipulation. Consequently, a representation or classifier that performs well on Earth may not transfer directly to reduced-gravity environments. This is particularly relevant for space robotics, where tactile perception systems must operate under conditions that are difficult to reproduce and validate experimentally on Earth. 

\begin{figure}[!t]
    \centering
    \includegraphics[width=0.9\linewidth]{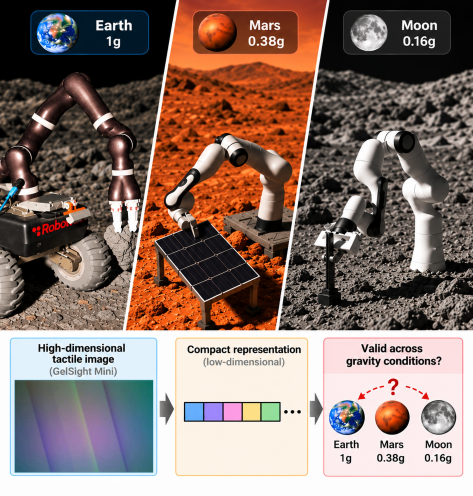}
    \caption{Multi-gravity scenarios considered in this work and overview of the investigated transfer problem.}
    \label{fig:scenario}
\end{figure}

Therefore, this work investigates two questions: (1)~whether a compact representation of tactile motion is sufficient to classify a manipulation event, and (2)~whether its performance transfers across gravity conditions. These questions are studied in simulation using the load-induced rotation of a grasped object as a controlled interaction event. The direction of rotation describes how the object moves relative to the grasp and can provide information for corrective manipulation before grasp stability is compromised. Unlike a wrist-mounted force/torque sensor, tactile sensing observes this motion directly at the contact interface and preserves its local spatial structure.

The tactile response is represented by dense optical flow between consecutive frames of a simulated GelSight Mini. The flow field is spatially aggregated over a fixed grid and used as input to lightweight classifiers, without providing gravity as an input. The same interaction task and representation are evaluated under Earth, Mars, Moon, and orbital gravity. This controlled setup allows us to measure how gravity affects the tactile representation and whether this domain shift affects classification performance (Fig.~\ref{fig:scenario}).

The analysis first quantifies the effect of gravity on the optical-flow representation. Classifiers trained only under Earth gravity are then evaluated on unseen gravity domains to assess cross-gravity generalization. Finally, a single model is trained using data from all four gravity conditions, followed by an ablation study to determine how far the tactile representation can be reduced while preserving classification performance.

The main contributions are:
\begin{itemize}
     \item A quantitative analysis of the gravity-induced domain shift in a compact tactile optical-flow representation, showing that classifiers trained only under Earth gravity lose accuracy under unseen reduced-gravity conditions.
    
     \item A cross-gravity classification approach based on a single model trained across Earth, Mars, Moon, and orbital gravity, achieving consistent performance across all four domains without requiring gravity as an input.
    
    \item A reduced optical-flow representation of 40 features that retains 95.7\% accuracy, less than one percentage point below the full 126-feature representation.

\end{itemize}

Together, these contributions provide a basis for developing computationally efficient tactile perception methods that account for gravity-domain shifts in space robotic manipulation.

%%%%%%%%%%%%%%%%%%%%%%%%%%%%%%%%%%%%%%%%%%%%%%%%%%%%%%%%%%%%%%%%%%%%%%%%%%%%%%%%
\section{Related Work}
\label{sec:related}
\subsection{Vision-Based Tactile Sensing}
Vision-based tactile sensors convert contact into images, either marker-based or intensity-based \cite{Li2025VBTSReview}, and have become a primary route to high-resolution touch for manipulation \cite{Yamaguchi2019Tactile}. Sensors of this type measure the deformation of a soft elastomer with a camera and infer geometry, force, and slip from it \cite{Yuan2017GelSight}. Force-oriented reviews show that marker tracking, optical flow, physical modelling, and learning-based methods are all used to estimate contact forces \cite{Fang2025ForceReview}, and recent learned estimators regress force distributions or 3D forces directly from vision-based tactile images \cite{Helmut2025FEATS}. Optical flow in particular has been used to estimate contact shape and force in DelTact \cite{Zhang2022DelTact:Pattern}, force, geometry, and depth in GelFlow \cite{Zhang2023GelFlow}, and 3D contact point clouds \cite{Du2022TactileFlow3D}. In contrast, this work evaluates a compact optical-flow representation of consecutive markerless tactile images for classifying the direction of load-induced rotation and analyses its transferability across gravity conditions.

\subsection{Tactile Simulation}
Tactile simulation enables controlled, reproducible datasets without real hardware. TACTO \cite{Wang2022TACTO}, Taxim \cite{Si2022Taxim}, and TacSL \cite{Akinola2025TacSL:Learning} render vision-based tactile images from simulated contact, and SimTacLS extends the idea to large-scale tactile skin \cite{Luu2023SimTacLS}. Models trained on simulated tactile images do not transfer to real sensors by default, and the gap is addressed with physically accurate simulation \cite{Sferrazza2020SimToReal} or domain adaptation \cite{Jing2023DomainAdaptation}. The present study uses TacEx, which combines contact simulation with tactile rendering in Isaac Sim for GelSight Mini \cite{Nguyen2024TacEx}, and it treats gravity as the domain variable of interest rather than the sim-to-real gap. Only the grayscale version of the simulated images is used, in the PhysX rigid-body configuration.

\subsection{Tactile Sensing in Space Robotics}
Space robotic manipulation involves contact-rich tasks such as capture, servicing, assembly, and sample collection, where contact dynamics and ground verification remain major challenges under microgravity and reduced gravity \cite{Papadopoulos2021SpaceSurvey}. Tactile sensing has entered this domain primarily through hardware designed to survive the environment. Jahanshahi and Zhu review resistive, capacitive, piezoelectric, and optical technologies for orbital and planetary platforms \cite{Jahanshahi2025SpaceTactile}, and the same authors review learning-based grasp control for space \cite{Jahanshahi2024MLReview}. Fibre-optic tactile pads have been demonstrated on a gripper designed for the Astrobee free-flying robots \cite{Frishman2021FBG}, a sensor based on optical-fibre knots measures pressure, hardness, and texture for on-orbit servicing \cite{Yu2023Optum}, and force-sensing grippers have been proposed for on-orbit capture \cite{Wu2025OrbitGripper}. Contact experiments have also flown, with gecko-inspired adhesive grippers tested aboard the International Space Station \cite{Chen2022Astrobee}. That gravity changes the control of a grasp is documented in human studies, where the coordination of grip and load forces adapts differently under Mars, Moon, and microgravity during parabolic flight \cite{Opsomer2018Gravity}. The effect of gravity on the tactile signal itself, on the features that a classifier consumes, has not been measured in any of these works. This study addresses that gap with one representation and one controlled task, so that the observed change can be attributed to gravity, and asks whether rotation-direction classification generalises across gravities without richer tactile images, marker displacement, reconstructed geometry, learned force maps, robot state, or a gravity input.

%%%%%%%%%%%%%%%%%%%%%%%%%%%%%%%%%%%%%%%%%%%%%%%%%%%%%%%%%%%%%%%%%%%%%%%%%%%%%%%%
\section{Tactile Representation and Classification} %Tactile Processing Pipeline
\label{sec:method}

\subsection{Strategy Overview}
\begin{figure*}[t]
 \centering
 \includegraphics[width=0.98\textwidth]{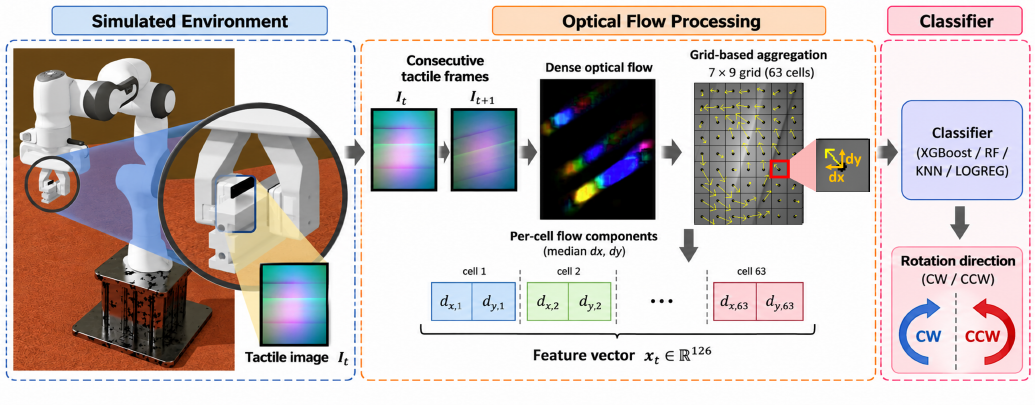}
 \caption{Processing pipeline from simulated tactile acquisition to rotation-direction classification. Consecutive tactile frames are converted to dense optical flow, spatially aggregated over a fixed grid, and used to classify load-induced rotation as clockwise (cw) or counterclockwise (ccw).}
 \label{fig:pipeline}
\end{figure*}

The processing pipeline consists of three stages (Fig.~\ref{fig:pipeline}).
First, a simulated vision-based tactile sensor records the contact
response while a load is applied to the grasped object under four
gravity conditions. Second, consecutive tactile frames are converted
to dense optical flow and spatially aggregated into a 126-feature
representation. Finally, lightweight classifiers use this
representation to predict the direction of load-induced rotation.

Two training regimes are considered. The first uses only Earth-gravity
data to evaluate transfer to unseen gravity domains, while the second
uses data from all four gravity conditions. A subsequent feature
ablation evaluates how far the representation can be reduced while
preserving classification performance. Gravity is not provided as an
input to the classifiers.

\subsection{Optical Flow Representation}
\label{sec:representation}
The input consists of simulated tactile image sequences of one sensor, converted to grayscale. The relevant cue is the motion of the
contact pattern between consecutive frames, which can be represented
from image intensity alone. 

For each pair of consecutive frames \(I_{t-1}\) and \(I_t\), dense optical flow is computed with the classical Farneb\"{a}ck polynomial-expansion method \cite{Farneback2003TwoFrame} implemented in OpenCV 4.6.0:

\begin{equation}
F_t = \mathrm{Flow}(I_{t-1}, I_t).
\end{equation}

A fixed configuration is used throughout the experiments: pyramid
scale 0.5, 3 levels, window size 25, 5 iterations, polynomial
neighbourhood 5, polynomial sigma 1.2, and default flags.

The resulting dense field is spatially reduced to a fixed grid.
Each \(240\times320\) px image is divided into \(R=7\) rows and
\(C=9\) columns, producing 63 cells. Within each cell, the median
horizontal and vertical flow components are retained as \(dx\) and
\(dy\), respectively. The resulting representation is

\begin{equation}
\mathbf{x}_t = \big[dx_{1},\ldots,dx_{63},\, dy_{1},\ldots,dy_{63}\big] \in \mathbb{R}^{126}\\
\end{equation}

\subsection{Rotation-Direction Target}
\label{sec:target}
The label of an event is the direction of rotation of its contact pattern, \texttt{cw} or \texttt{ccw}, and every flow field of the event carries that label. Because the two pads of the gripper face the object from opposite sides, the same load produces opposite rotation directions on the two sides (Table~\ref{tab:label_map}). Each sensor is therefore treated as an
independent sample, while data from both sensors are used to train the
classifier.

Only the optical-flow representation is provided to the classifier.
Force magnitude, load position, gravity condition, repetition, sensor
side, and frame index are not included as input features.

\begin{table}[t]
\centering
\caption{Rotation-direction label as a function of load direction and sensor side.}
\label{tab:label_map}
\begin{tabular}{lcc}
\toprule
Load & Left sensor & Right sensor \\
\midrule
Downward $F_z$ & ccw & cw \\
Upward $F_z$ & cw & ccw \\
\bottomrule
\end{tabular}
\end{table}

\subsection{Feature Selection}
\label{sec:selection}

The 126-dimensional representation contains the horizontal and
vertical flow medians of the 63 grid cells. Three feature
configurations are evaluated. Configuration A uses the full
representation. Configuration B uses PCA-guided selection:
principal component analysis \cite{Jolliffe2002PCA} identifies the
components explaining 90\% of the variance in the training data, and
the original features are ranked according to their contribution to
these components. Configuration C uses minimum redundancy maximum
relevance (mRMR) \cite{Peng2005mRMR}, which ranks features by their
relevance to the target while penalising redundancy with previously
selected features. The ranking is truncated at its elbow.

Both selection methods retain subsets of the original flow features;
they do not transform the classifier input. Feature selection is
performed using the training set only. In addition, a grid-resolution
ablation evaluates increasingly compact spatial representations and
the contribution of the individual flow components.

\subsection{Classifiers and Training}
\label{sec:training}
Four lightweight classifiers are evaluated: XGBoost, random forest,
k-nearest neighbours (KNN), and logistic regression. KNN uses
standardised features, while logistic regression provides a linear
reference. These models are suitable for compact hand-crafted tactile
representations and have been used for related tactile classification
tasks \cite{Elijah2025SlipModelSelection}.

The dataset is divided into 80\% training and 20\% test data. Splitting
is performed at the trial level and stratified by class, ensuring that
events from the same trial remain in the same partition. This prevents
information from the same trial from appearing in both training and
test sets.

Two training regimes are evaluated. In the \emph{Earth} regime, only
Earth-gravity data are used for training, while the test data are
separated by gravity to evaluate Earth performance and transfer to the
unseen Mars, Moon, and orbital domains. In the \emph{All} regime,
training data include all four gravity conditions. Hyperparameters for XGBoost, random forest, and KNN are selected from
the training set using grouped 3-fold cross-validation with Bayesian
optimisation (Optuna, 12 trials, accuracy criterion). Logistic
regression uses fixed hyperparameters. Each final model is fitted to
the complete training set and evaluated once on the held-out test set. Accuracy per flow field is the primary metric. Balanced accuracy,
macro-F1, ROC-AUC, and PR-AUC \cite{Davis2006PRROC} are also
reported.

%%%%%%%%%%%%%%%%%%%%%%%%%%%%%%%%%%%%%%%%%%%%%%%%%%%%%%%%%%%%%%%%%%%%%%%%%%%%%%%%
\section{Experimental Setup}
\label{sec:setup}

\subsection{Simulation and Load Application}
The experiments are conducted in the Space Robotics Bench (SRB),
built on Isaac Sim for the development and validation of robotic
systems in space scenarios \cite{Orsula2025SRB}. Vision-based
tactile sensing is simulated through TacEx
\cite{Nguyen2024TacEx} using its PhysX rigid-body
configuration. In this configuration, the gel pad is represented as
a rigid body with compliant contacts. This configuration provides
stable rendering of contact-pattern rotation over the large number
of trials required in this study, while avoiding additional
deformation effects from a simulated soft gel
(Section~\ref{sec:limitations}).
 
\begin{figure}[t]
    \centering
    \includegraphics[width=1.0\linewidth]{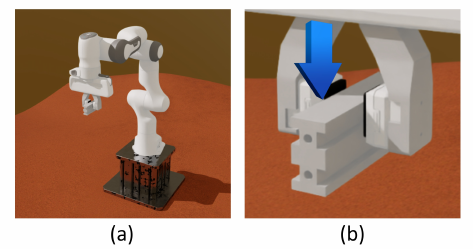}
    \caption{The Space Robotics Bench simulation environment. (a) Gripper with integrated GelSight Mini in a Martian environment. (b) Applied force.}
    \label{fig:simulation_setup}
\end{figure}

The simulated setup (Fig.~\ref{fig:simulation_setup}) consists of a
Franka arm with two vision-based tactile sensors mounted on a parallel
gripper. A metallic profile is grasped horizontally at its centre
between the two sensors. During each trial, the manipulator maintains
a fixed grasp while an external vertical force \(F_z\) is applied to
the profile at a distance \(d\) from the grasp, producing a torque
about an axis parallel to the sensor optical axis.

Three force magnitudes (3, 4.5, and 6\,N) are applied upward or
downward for 2\,s at 20 positions corresponding to moment arms from
75 to 100\,mm. Varying force magnitude and moment arm exposes the
classifier to a range of torques rather than a single loading
condition. The complete experimental design comprises
3 force magnitudes \(\times\) 2 directions \(\times\) 20 positions
\(\times\) 4 repetitions \(\times\) 4 gravity conditions:
Earth, Mars, Moon, and orbital gravity (0g). The simulator is fully
reset between trials.

\subsection{Dataset and Partitions}

Each recording from one tactile sensor during one load application is
treated as an event, giving two events per trial. A total of 3840
events were recorded. The recordings were reviewed to retain events
with measurable contact-pattern rotation and a consistent rotation
direction. Events dominated by linear slip, including cases in which
the profile slid out of the grasp without rotating, were excluded
because linear slip is outside the classification target.

The resulting dataset contains 2366 events from 1245 trials, with
563--620 events per gravity condition. The two rotation classes are
approximately balanced, with 1232 \texttt{ccw} and 1134 \texttt{cw}
events. Optical-flow fields are extracted from the interval in which
the contact pattern rotates, and each field inherits the label of its
event. Table~\ref{tab:partition} summarizes the partitions used in the
two training regimes.

\begin{table}[t]
\centering
\footnotesize
\caption{Partitions of the two training regimes: number of events and of flow fields in each set.}
\label{tab:partition}
\setlength{\tabcolsep}{4pt}
\begin{tabular}{llrr}
\toprule
Regime & Set & Events & Flow fields \\
\midrule
All & Train (four gravities) & 1893 & 37199 \\
All & Test set (four gravities) & 473 & 9403 \\
\midrule
Earth & Train (Earth) & 468 & 8935 \\
Earth & Test, Earth & 118 & 2469 \\
Earth & Test, Mars & 120 & 2336 \\
Earth & Test, Moon & 114 & 2242 \\
Earth & Test, Orbit (0g) & 123 & 2439 \\
\bottomrule
\end{tabular}
\end{table}

%%%%%%%%%%%%%%%%%%%%%%%%%%%%%%%%%%%%%%%%%%%%%%%%%%%%%%%%%%%%%%%%%%%%%%%%%%%%%%%%
\section{Results}
\label{sec:experiments}

\subsection{Gravity-Domain Analysis}
\label{sec:gravity}

The first analysis tests whether gravity changes the optical-flow representation before considering classification performance. The PERMANOVA \cite{Anderson2001PERMANOVA} measures the joint effect of gravity without assuming any distribution of the features, the Kruskal--Wallis tests \cite{Kruskal1952} locate it feature by feature, and the False Discovery Rate (FDR) correction \cite{Benjamini1995FDR} accounts for the 126 comparisons. 

The PERMANOVA shows a statistically significant but small effect of gravity
(\(R^2=0.0164\), pseudo-\(F=13.12\), \(p=0.001\), 999 permutations).
Gravity therefore accounts for approximately 1.6\% of the variance in
the representation. Its effect is smaller than those of load position
(\(R^2=0.591\)), sensor side (0.189), force magnitude (0.055), and
load direction (0.043).

\begin{figure}[t]
\centering
\includegraphics[width=\linewidth]{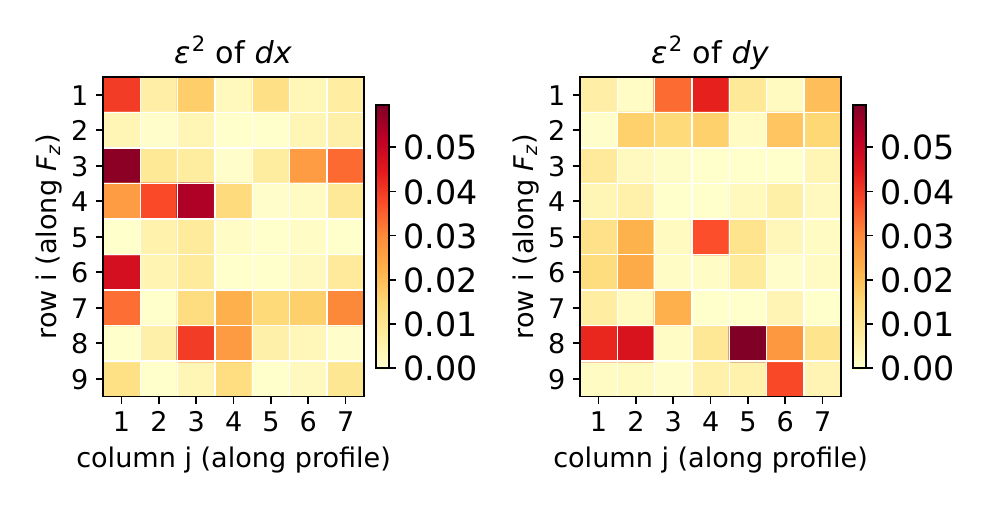}
\caption{Gravity sensitivity per cell (Kruskal--Wallis effect size \(\epsilon^2\)) for the \(dx\) and \(dy\) components, in the world frame (columns along the profile, rows along \(F_z\)).}
\label{fig:eps2_map}
\end{figure}

Pairwise tests with FDR correction distinguish Earth from Mars, Moon,
and orbit (\(R^2=0.017\)--0.020, \(p=0.002\)). Mars is also
distinguishable from Moon and orbit (\(R^2=0.002\)--0.003,
\(p=0.002\)), whereas Moon and orbit are not
(\(R^2=0.0008\), \(p=0.53\)). At the feature level, the Kruskal--Wallis tests identify significant gravity effects in 78 of the 126 features, distributed between flow components and multiple grid cells (Fig.~\ref{fig:eps2_map}). The largest effect
sizes are \(\epsilon^2=0.0595\) for \(dy\) and 0.0579 for \(dx\).

Overall, gravity introduces a small but measurable shift
in the optical-flow representation. The following analysis evaluates whether this shift affects classifiers trained only under Earth gravity.

\subsection{Earth-Only Training}
\label{sec:earth}
\begin{figure}[t]
\centering
\includegraphics[width=0.85\linewidth]{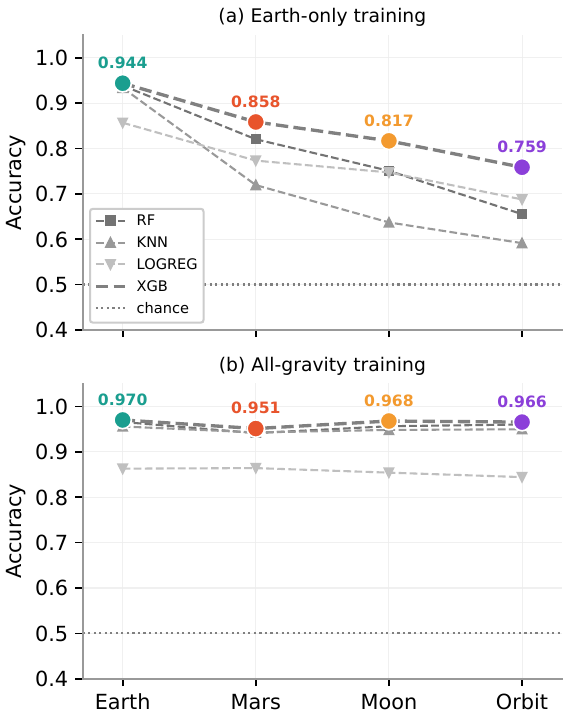}
\caption{Accuracy per flow field and gravity domain for the four classifiers (XGBoost highlighted). (a) Earth-only training, where Mars, Moon, and orbit are unseen domains. (b) All-gravity training.}
\label{fig:e2_degradation}
\end{figure}

The Earth-only regime evaluates whether a classifier trained under
Earth gravity transfers to unseen gravity domains. XGBoost achieves
0.944 accuracy on Earth but decreases to 0.858 on Mars, 0.817 on the
Moon, and 0.759 in orbit (Table~\ref{tab:e2_earth} and
Fig.~\ref{fig:e2_degradation}(a)). This corresponds to a maximum
accuracy decrease of 0.185.

The same trend appears in all four classifiers. From Earth to orbit, random forest decreases from 0.938 to 0.655, KNN from 0.934 to 0.591, and logistic regression from 0.857 to 0.688. For every classifier, accuracy decreases progressively from Earth to Mars, Moon, and orbit. Balanced accuracy and macro-F1 show the same pattern,
as do ROC-AUC and PR-AUC. These results show that the gravity-domain shift identified in the optical-flow features is accompanied by reduced classification performance when training is restricted to Earth-gravity data.

\begin{table*}[t]
\centering
\footnotesize
\caption{Earth-only training: metrics per flow field on the Earth test events and on the three unseen gravity domains, per model, with the accuracy drop of each domain relative to Earth.}
\label{tab:e2_earth}
\setlength{\tabcolsep}{1.5pt}
\begin{tabular*}{\textwidth}{@{\extracolsep{\fill}}lcccccccccccccccc@{}}
\toprule
Metric & \multicolumn{4}{c}{\textbf{XGB}} & \multicolumn{4}{c}{RF} & \multicolumn{4}{c}{KNN} & \multicolumn{4}{c}{LOGREG} \\
\cmidrule(lr){2-5}\cmidrule(lr){6-9}\cmidrule(lr){10-13}\cmidrule(lr){14-17}
 & Earth & Mars & Moon & Orbit & Earth & Mars & Moon & Orbit & Earth & Mars & Moon & Orbit & Earth & Mars & Moon & Orbit \\
\midrule
\textbf{Accuracy} & \textbf{0.944} & \textbf{0.858} & \textbf{0.817} & \textbf{0.759} & \textbf{0.938} & \textbf{0.820} & \textbf{0.751} & \textbf{0.655} & \textbf{0.934} & \textbf{0.719} & \textbf{0.637} & \textbf{0.591} & \textbf{0.857} & \textbf{0.773} & \textbf{0.747} & \textbf{0.688} \\
Balanced acc. & 0.944 & 0.858 & 0.817 & 0.760 & 0.939 & 0.822 & 0.759 & 0.665 & 0.933 & 0.720 & 0.644 & 0.596 & 0.856 & 0.772 & 0.746 & 0.688 \\
Macro-F1 & 0.944 & 0.858 & 0.816 & 0.758 & 0.938 & 0.819 & 0.748 & 0.648 & 0.934 & 0.718 & 0.634 & 0.590 & 0.856 & 0.772 & 0.746 & 0.687 \\
ROC-AUC & 0.991 & 0.932 & 0.898 & 0.842 & 0.989 & 0.929 & 0.877 & 0.782 & 0.964 & 0.773 & 0.693 & 0.630 & 0.958 & 0.860 & 0.843 & 0.786 \\
PR-AUC & 0.991 & 0.939 & 0.897 & 0.843 & 0.989 & 0.933 & 0.877 & 0.769 & 0.949 & 0.719 & 0.622 & 0.580 & 0.958 & 0.870 & 0.838 & 0.785 \\
\midrule
Accuracy drop vs.\ Earth & -- & 0.085 & 0.127 & 0.185 & -- & 0.118 & 0.188 & 0.283 & -- & 0.214 & 0.297 & 0.342 & -- & 0.084 & 0.110 & 0.169 \\
\bottomrule
\end{tabular*}
\end{table*}

 \subsection{All-Gravity Training}
\label{sec:all}

When training data from all four gravity domains are included,
XGBoost achieves 0.963 accuracy on the held-out test set
(Table~\ref{tab:e1_all}). Random forest and KNN reach 0.955 and
0.949, respectively, while logistic regression reaches 0.857.
XGBoost also achieves a ROC-AUC and PR-AUC of 0.996.

Performance remains similar across the individual gravity domains
despite gravity not being provided as an input to the classifier
(Table~\ref{tab:e1_break} and Fig.~\ref{fig:e2_degradation}(b)).
XGBoost accuracy ranges from 0.951 on Mars to 0.970 on Earth, with
0.968 on the Moon and 0.966 in orbit. Similar performance is also
observed across load magnitudes, sensor sides, and rotation
magnitudes.

Thus, representing all four gravity domains during training yields
consistent classification performance across the conditions evaluated
in this study.

\begin{table}[!h]
\centering
\footnotesize
\caption{All-gravity training: metrics per flow field of the four classifiers on the test set.}
\label{tab:e1_all}
\setlength{\tabcolsep}{2pt}
\begin{tabular}{lccccc}
\toprule
Model & \textbf{Accuracy} & Bal.\ acc. & Macro-F1 & ROC-AUC & PR-AUC \\
\midrule
\textbf{XGB} & \textbf{0.963} & 0.963 & 0.963 & 0.996 & 0.996 \\
RF & \textbf{0.955} & 0.955 & 0.955 & 0.994 & 0.994 \\
KNN & \textbf{0.949} & 0.948 & 0.948 & 0.976 & 0.965 \\
LOGREG & \textbf{0.857} & 0.855 & 0.856 & 0.952 & 0.953 \\
\bottomrule
\end{tabular}
\end{table}

\begin{table*}[t]
\centering
\small
\caption{All-gravity training, XGBoost breakdowns by the factors excluded from the input: metrics per flow field on the test set.}
\label{tab:e1_break}
\begin{tabular*}{\textwidth}{@{\extracolsep{\fill}}lcccccccccccc@{}}
\toprule
Metric & \multicolumn{4}{c}{By gravity} & \multicolumn{3}{c}{By load magnitude} & \multicolumn{2}{c}{By sensor side} & \multicolumn{3}{c}{By rotation magnitude} \\
\cmidrule(lr){2-5}\cmidrule(lr){6-8}\cmidrule(lr){9-10}\cmidrule(lr){11-13}
 & Earth & Mars & Moon & Orbit & 3.0 N & 4.5 N & 6.0 N & Left & Right & $<5^\circ$ & 5 to $15^\circ$ & $>15^\circ$ \\
\midrule
\textbf{Accuracy} & \textbf{0.970} & \textbf{0.951} & \textbf{0.968} & \textbf{0.966} & \textbf{0.968} & \textbf{0.955} & \textbf{0.975} & \textbf{0.967} & \textbf{0.959} & \textbf{0.960} & \textbf{0.960} & \textbf{0.974} \\
Balanced acc. & 0.970 & 0.952 & 0.968 & 0.966 & 0.968 & 0.955 & 0.975 & 0.957 & 0.953 & 0.962 & 0.968 & 0.974 \\
Macro-F1 & 0.970 & 0.951 & 0.968 & 0.966 & 0.967 & 0.955 & 0.975 & 0.962 & 0.956 & 0.958 & 0.940 & 0.974 \\
ROC-AUC & 0.997 & 0.994 & 0.997 & 0.996 & 0.997 & 0.994 & 0.998 & 0.995 & 0.995 & 0.995 & 0.996 & 0.998 \\
PR-AUC & 0.997 & 0.993 & 0.996 & 0.996 & 0.996 & 0.994 & 0.998 & 0.991 & 0.997 & 0.997 & 0.984 & 0.998 \\
\bottomrule
\end{tabular*}
\end{table*}

\subsection{Minimal Representation}
\label{sec:minimal}

\begin{figure}[t]
\centering
\includegraphics[width=0.85\linewidth]{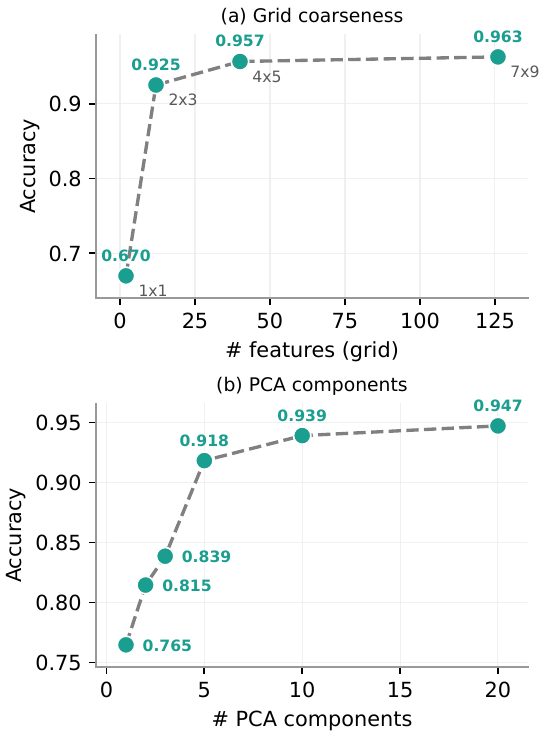}
\caption{Minimal representation: Accuracy versus feature count for grid resolution (a) and PCA (b).}
\label{fig:e3_curves}
\end{figure}

The ablation study evaluates how far the optical-flow representation
can be reduced while preserving classification performance. XGBoost
is evaluated with fixed hyperparameters and the same train/test
partition while varying the flow components, grid resolution, and
feature-selection strategy (Table~\ref{tab:e3_minimal} and
Fig.~\ref{fig:e3_curves}).

Reducing the spatial grid from \(7\times9\) (126 features) to
\(4\times5\) (40 features) decreases accuracy only from 0.963 to
0.957. PCA-guided selection reaches the same accuracy with 74
features. More aggressive reductions produce larger losses: mRMR
selection reaches 0.939 with 14 features, a \(2\times3\) grid reaches
0.925 with 12 features, and a single-cell representation reaches
0.670 with two features. Using only one flow component also reduces
accuracy, to 0.944 for \(dx\) and 0.947 for \(dy\).

The \(4\times5\) grid therefore provides the most compact spatial
representation evaluated here while remaining within one percentage
point of the full 126-feature representation.

\begin{table*}[t]
\centering
\small
\caption{Minimal representation (XGBoost, fixed hyperparameters): metrics per flow field on the test set against the number of features.}
\label{tab:e3_minimal}
\begin{tabular}{lcccccc}
\toprule
Representation & \# features & \textbf{Accuracy} & Balanced acc. & Macro-F1 & ROC-AUC & PR-AUC \\
\midrule
$dx$ only & 63 & \textbf{0.944} & 0.944 & 0.944 & 0.991 & 0.991 \\
$dy$ only & 63 & \textbf{0.947} & 0.947 & 0.947 & 0.991 & 0.991 \\
$dx+dy$ (full) & 126 & \textbf{0.963} & 0.963 & 0.963 & 0.996 & 0.996 \\
\midrule
Grid 1$\times$1 & 2 & \textbf{0.670} & 0.669 & 0.668 & 0.727 & 0.729 \\
Grid 2$\times$3 & 12 & \textbf{0.925} & 0.925 & 0.925 & 0.982 & 0.982 \\
Grid 4$\times$5 & 40 & \textbf{0.957} & 0.957 & 0.957 & 0.994 & 0.994 \\
Grid 7$\times$9 & 126 & \textbf{0.963} & 0.963 & 0.963 & 0.996 & 0.996 \\
\midrule
PCA-guided 90\% (B) & 74 & \textbf{0.957} & 0.957 & 0.957 & 0.994 & 0.994 \\
mRMR elbow (C) & 14 & \textbf{0.939} & 0.939 & 0.939 & 0.988 & 0.988 \\
\bottomrule
\end{tabular}
\end{table*}

\subsection{Computational Cost}
\label{sec:cost}
Runtime is measured on a single thread of an Intel Core Ultra 7 268V
under Linux. The reported values are therefore machine-dependent and
are intended to characterize the relative computational cost of the
pipeline and classifiers.

Feature extraction dominates inference time. Farneb\"{a}ck optical
flow on a \(320\times240\) frame requires 10.15\,ms, followed by
1.43\,ms for the \(7\times9\) grid aggregation. In comparison,
XGBoost classification requires only 0.14\,ms per flow field
(Table~\ref{tab:cost}). Logistic regression requires 0.11\,ms, KNN
2.19\,ms, and random forest 33.62\,ms.

XGBoost also provides the highest classification accuracy in both
training regimes while requiring a 0.4\,MB fitted model and 2.07\,s
for a single training fit. For this experimental configuration,
classification therefore contributes little to the total processing
time; dense optical-flow computation is the main computational
bottleneck.

\begin{table}[t]
\centering
\small
\caption{Classifier cost on one CPU thread: training time (median of 5 fits), inference time per flow field (median of 200 runs), and model size.}
\label{tab:cost}
\setlength{\tabcolsep}{3pt}
\begin{tabular}{lccc}
\toprule
Model & Training (s) & Inference (ms) & Model size (MB) \\
\midrule
\textbf{XGB} & 2.07 & 0.14 & 0.4 \\
RF & 144 & 33.62 & 58.9 \\
KNN & 0.03 & 2.19 & 19.1 \\
LOGREG & 0.21 & 0.11 & 0.005 \\
\bottomrule
\end{tabular}
\end{table}

%%%%%%%%%%%%%%%%%%%%%%%%%%%%%%%%%%%%%%%%%%%%%%%%%%%%%%%%%%%%%%%%%%%%%%%%%%%%%%%%
\section{Discussion}
\label{sec:discussion}

\subsection{Gravity-Induced Domain Shift}

Gravity has a small effect on the overall variability of the
optical-flow representation but a substantial impact on cross-gravity
classification. The PERMANOVA attributes only 1.6\% of the feature
variance to gravity, considerably less than to load position, sensor
side, force magnitude, or load direction. Nevertheless, under
Earth-only training, XGBoost accuracy decreases from 0.944 on Earth
to 0.759 in orbit, with the same degradation trend observed across
all four classifiers.

These results are not contradictory: a small contribution to the
total feature variance can still affect the feature structure relevant
to the classification boundary. Moreover, the degradation increases
from Mars to Moon and orbit as gravity departs from Earth's. The two
analyses therefore provide complementary evidence: gravity produces
a small but measurable shift in the tactile representation, and this
shift is associated with substantial performance degradation when
classifiers trained on Earth are transferred to reduced-gravity
domains.

\subsection{Generalisation Across Gravity Domains}
Including all four gravity domains in training results in consistent performance across them. XGBoost achieves 0.963 overall accuracy,
with per-gravity accuracy ranging from 0.951 to 0.970, despite gravity
not being provided as an input. This shows that a single classifier
can operate across the four evaluated gravity domains when they are
represented during training.

The Earth-only results also show that sensitivity to the domain shift
depends on the classifier. The maximum accuracy loss ranges from 0.169
for logistic regression to 0.342 for KNN, while XGBoost combines the
highest accuracy with greater robustness to the shift. These results
indicate that cross-gravity performance depends on both the tactile
representation and the classifier used to interpret it.

\subsection{Minimal Representation and Cost}
The representation can be substantially reduced with little loss in
accuracy. A \(4\times5\) grid reduces the feature set from 126 to 40
while decreasing accuracy only from 0.963 to 0.957. More aggressive
compression produces a clearer trade-off: mRMR retains 0.939 accuracy
with 14 features, whereas a single-cell representation falls to 0.670.

The main computational cost, however, remains in feature extraction.
Farneb\"{a}ck optical flow and grid aggregation require approximately
11.6\,ms per frame pair, compared with only 0.14\,ms for XGBoost
inference. A coarser grid therefore reduces the representation size,
but further runtime improvements would require reducing the cost of
the optical-flow computation itself.

\subsection{Limitations}
\label{sec:limitations}

Two limitations bound these findings. First, the study is
simulation-based and uses a rigid gel model. A deformable gel may produce different optical-flow responses, so validation with a real tactile sensor is needed. Second, the classifier is sensitive to sensor orientation. Deployment with different orientations would therefore require a consistent image reference frame or orientation-diverse training data.

%%%%%%%%%%%%%%%%%%%%%%%%%%%%%%%%%%%%%%%%%%%%%%%%%%%%%%%%%%%%%%%%%%%%%%%%%%%%%%%%
\section{Conclusion and Future Work}
\label{sec:conclusion}
This simulation study investigated whether a compact optical-flow representation can classify load-induced rotation and transfer across gravity domains. The 126-feature representation achieves 0.963 accuracy, while a reduced 40-feature grid retains 0.957. Although gravity accounts for only 1.6\% of the feature variance, classifiers
trained only on Earth data degrade under reduced gravity, with XGBoost accuracy decreasing from 0.944 on Earth to 0.759 in orbit. In contrast, a single model trained across all four gravity domains achieves consistent performance in each domain without using gravity as an input. XGBoost inference requires only 0.14\,ms, with dense optical-flow computation remaining the main computational cost.

These results show that Earth-gravity performance alone is
insufficient to establish the transferability of tactile perception for space robotic manipulation. They also show that a compact representation can preserve most of the classification performance across the evaluated gravity conditions. Future work will focus on real-sensor validation, transfer to unseen gravity conditions, and
extension to a broader range of contact events, loading conditions, and object geometries.

\section{Acknowledgments}
This work was supported by the Luxembourg National Research Fund (AFR grant No. 19479896), in partnership with the SpaceR research group at SnT, University of Luxembourg, and by the Catalan Government through ACCIO-Eurecat (Traça-SESAM project). The views expressed are solely those of the authors and do not necessarily reflect those of the funding institutions.
%%%%%%%%%%%%%%%%%%%%%%%%%%%%%%%%%%%%%%%%%%%%%%%%%%%%%%%%%%%%%%%%%%%%%%%%%%%%%%%%
\bibliography{references_v2}

\end{document}